\documentclass{fcs}
\usepackage{bm}

\volumn{ }
\doi{ }
\articletype{RESEARCH~ARTICLE}
\copynote{{\copyright} Higher Education Press and Springer-Verlag Berlin Heidelberg 2012}
\ratime{Received month dd, yyyy; accepted month dd, yyyy}
\email{$ymh@nenu.edu.cn$ \\ Authors are listed alphabetically by last name.}
\title{PBCounter: Weighted Model Counting on Pseudo-Boolean Formulas}
\author{Yong LAI $^{1,2}$, Zhenghang XU $^{1,2}$, Minghao YIN\xff $^{2,3}$}
\address{
{1\quad College of Computer Science and Technology, Jilin University, Changchun 130012, China} \\
{2\quad Key Laboratory of Symbolic Computation and Knowledge Engineering Ministry of Education, Jilin University, Changchun 130012, China}\\
{3\quad School of Computer Science and Information Technology, Northeast Normal University, Changchun 130017, China} \\
}

\newcommand{\PBCounter}{\ensuremath{\mathsf{PBCounter}}}

\usepackage{amsfonts}
\usepackage{amssymb}
\usepackage{color}
\usepackage[linesnumbered,ruled,vlined]{algorithm2e} 
\usepackage{setspace}
\usepackage{enumerate}
\usepackage{graphics}
\usepackage{multirow}
\usepackage{booktabs}
\usepackage{caption}
\usepackage{tabularx}
\usepackage{soul}

\soulregister\cite7
\soulregister\ref7

\newtheorem{example}{Example}
\newtheorem{theorem}{Theorem}
\newtheorem{definition}{Definition}

\newcolumntype{Z}{>{\centering\let\newline\\\arraybackslash\hspace{0pt}}X}

\begin{document}
\maketitle
\setcounter{page}{1}
\setlength{\baselineskip}{14pt}

 \begin{abstract}
    In Weighted Model Counting (WMC), we assign weights to literals and compute the sum of the weights of the models of a given propositional formula where the weight of an assignment is the product of the weights of its literals.
    The current WMC solvers work on Conjunctive Normal Form (CNF) formulas. However, CNF is not a natural representation for human-being in many applications.
    Motivated by the stronger expressive power of pseudo-Boolean (PB) formulas than CNF, we propose to perform WMC on PB formulas.
    Based on a recent dynamic programming algorithm framework called ADDMC for WMC, we implement a weighted PB counting tool \PBCounter{}.
    We compare \PBCounter{} with the state-of-the-art weighted model counters SharpSAT-TD, ExactMC, D4, and ADDMC, where the latter tools work on CNF with encoding methods that convert PB constraints into a CNF formula.
    The experiments on three domains of benchmarks show that \PBCounter{} is superior to the model counters on CNF formulas.
\end{abstract}


\section{Introduction}

Model counting is a fundamental problem in computer science with a wide variety of applications in practice, ranging from neural network verification~\cite{Baluta:etal:19}, network reliability~\cite{Duenas-Osorio:etal:17}, computational biology~\cite{Sashittal:El-Kebir:19}, and the like.
Given a propositional formula $\varphi$, the model counting problem is to compute the number of satisfying assignments of $\varphi$. 
Weighted Model Counting (WMC) is a generalization of model counting.
In WMC, we assign weights to literals and compute the sum of the weights of the models of $\varphi$ where the weight of an assignment is the product of the weights of its literals.
The WMC solvers can serve as efficient tools for probabilistic inference~\cite{Roth:96,DBLP:conf/aaai/SangBK05,Chavira:Darwiche:08}, and many applications benefit significantly from WMC-based probabilistic inference, including
probabilistic planning~\cite{DBLP:journals/jair/DomshlakH07}, probabilistic databases~\cite{VdB:Suciu:17}, and probabilistic programming~\cite{Fierens:etal:15}. 

The major approaches to the design of scalable (exact) model counters are mainly divided into three types: (1) search-based, (2) compilation-based, and (3) variable elimination-based methods~\cite{DBLP:conf/aaai/LaiMY21}.
The {\it search-based} techniques predominately focus on the combination of component caching with Conflict Driven clause learning including Cachet~\cite{DBLP:conf/sat/SangBBKP04}, sharpSAT~\cite{DBLP:conf/sat/Thurley06} and Ganank~\cite{DBLP:conf/ijcai/SharmaRSM19}.
The {\it compilation-based} techniques rely on the paradigm of knowledge compilation. Target languages of interest include d-DNNF and CCDD, with two counters based on them, D4 and ExactMC~\cite{DBLP:conf/ijcai/LagniezM17,DBLP:conf/aaai/LaiMY21}.
Recently {\it variable elimination-based} techniques have been proposed, ADDMC uses the early project to eliminate variables represented by Algebraic Decision Diagrams~\cite{DBLP:conf/aaai/DudekPV20}. NestHDB is a scalable hybrid tool that combines search-based and variable elimination-based counting methods~\cite{hecher2020taming}.
There are also many approximate model counters, such as STS~\cite{ermon2012uniform}, ApproxMC~\cite{chakraborty2013scalable}, and PartialKC~\cite{lai2023fast}.

The current WMC solvers work on Conjunctive Normal Form (CNF) formulas. However, CNF is not a natural representation of human-beings in many applications.
A Pseudo-Boolean (PB) constraint is of the form $\sum_{i=1}^n a_il_i \vartriangleright b$, where the $a_i$ and $b$ are integers, literal $l_i$ is a Boolean variable $x_i$ or its negation, and the relation operator $\vartriangleright \in \{<, \leq, =, \geq, >\}$.
PB constraints are a natural generalization of clauses in CNF; e.g., a clause $l_1 \lor l_2 \lor \dots \lor l_m$ is equivalent to the PB constraint $\sum_{i=1}^m l_i \geq 1$. 
Compared with clauses, it is easier to use PB constraints to describe the arithmetic properties of a problem. 
Moreover, PB constraints are strictly more succinct than a CNF formula according to the criteria used in the knowledge compilation map~\cite{Le-Berre:etal:18}.
Thus, PB constraints have been widely used to efficiently represent many problems ranging from wireless sensor network~\cite{kovasznai2018investigations} and logic synthesis~\cite{DBLP:conf/iccad/AloulRMS02} to Minimum-Width Confidence Bands~\cite{DBLP:conf/sat/LeiCLH21}. 

Currently, a general approach to solving the problems modeled in PB constraints is to encode them into a CNF formula with the same satisfiability or model count and then employ a SAT solver or model counter to solve the resulting CNF formula. 
The clear benefit of this approach is due to the high-performance of the modern SAT solvers and model counters. 
However, the encoding from a PB formula to CNF introduces many auxiliary variables, and thus increases the size of the problem and even sometimes damages the structure or semantics of the original constraints.
For these reasons, developing tools that work directly on PB formulas makes sense.

For the satisfiability of PB constraints, there exist many complete tools that directly work on PB formulas, such as PBS~\cite{aloul2002pbs}, PUEBLO~\cite{DBLP:journals/jsat/SheiniS06}, and RoundingSat~\cite{Elffers:Nordstrom:18}.
These tools are more efficient on PB constraints since they use cutting planes proof system to analyze conflicts, which was shown exponentially more powerful than the resolution-based proof system in normal SAT solving.
There exist some incomplete tools for PB Optimization such as NuPBO~\cite{chu2023towards} and DeepOpt-PBO~\cite{zhou2023improving} as well.
Turning to model counting or WMC, there are no tools that directly work on PB formulas so far.
In this study, we propose to perform WMC on PB formulas.
Based on a recent dynamic programming algorithm framework called ADDMC for WMC on CNF formulas, we implement a weighted PB counting tool \PBCounter{}.
Our experiments have demonstrated that \PBCounter{} was able to solve the greatest number of instances and was the fastest on most of them.

Model counting on PB constraints can also be viewed as a counting problem for 0-1 integer linear programming, which is a simplified version of integer linear programming (ILP) that restricts the variables to take values of 0 or 1.
Counting integer solutions of ILP has important applications such as simple temporal planning~\cite{huang2018new} and probabilistic program analysis~\cite{luckow2014exact}. 
In practice, however, it is often difficult to count the solutions of an ILP problem with a great dimension~\cite{ge2019approximating}.
We compare \PBCounter{} with the ILP counter IntCount\cite{ge2021decomposition}, and the experimental results show that \PBCounter{} has a significant advantage.

The rest of the paper is organized as follows: We present notations and preliminaries in Section \ref{sec:preliminaries}. 
Our algorithm of \PBCounter{} is presented in Section \ref{sec:WMConPB}.
In Section \ref{sec:preTech}, we present two preprocessing techniques on PB Constraints.
Next, we present a detailed empirical evaluation in Section \ref{sec:ExpRes}. 
Finally, we conclude in Section \ref{sec:Conclusion}.

\section{Preliminaries}
\label{sec:preliminaries}

In this section, we introduce the background knowledge about Weighted Model Counting (WMC), Pseudo-Boolean (PB) constraints, and Algebraic Decision Diagram (ADD), the primary data structure we used in our tool \PBCounter{}. 

\subsection{Weighted Model Counting}

Given a set of Boolean variables $\{x_1, x_2 \ldots x_n\}$, a literal is either a variable $x_i$ or its negation $\overline {x_i}$.
The negation of a literal $l$, written $\overline l$, denotes $\overline x$ if $l$ is $x$, and $x$ if $l$ is $\overline x$.
A clause is a disjunction of literals. 
A conjunctive normal form (CNF) formula $F = c_1 \land c_2 \land \dots \land c_m$ is a conjunction of clauses.

\begin{definition}[Weighted Model Counting]
	\label{def:WMC}
	Given a CNF formula $\psi$ over variables set $X$, $\psi$ can be represented as a Boolean function $\psi : 2^{X} \to \{0, 1\}$. Let $W:2^{X} \to \mathbb{R}$ be an arbitrary function representing the weight of assignments. The weighted model count of $\psi$ w.r.t. $W$ is $W(\psi) = \sum_{\tau \in 2^{X}} \psi(\tau) \cdot W(\tau)$ .
\end{definition}

The function $W:2^{X} \to \mathbb{R}$ is called the weight function.
The weight of an assignment is the product of the weight of its literals.
Next, we define two operations that are useful for WMC solving.

\begin{definition} [Product]
	\label{def:product}
    Let $X$ and $Y$ be sets of variables. The product of functions $f:2^X\to \mathbb{R}$ and $g:2^Y \to \mathbb{R}$ is the function $f\cdot g :2^{X\cup Y} \to \mathbb{R}$ defined for all $\tau \in 2^{X\cup Y}$ by $(f\cdot g)(\tau) := f(\tau_X) \cdot g(\tau_Y)$ .
\end{definition}

In Definition \ref{def:product}, $\tau_X$ is the projection of $\tau$ over variable set $X$.
Operation $product$ is commutative and associative.
If we use Boolean function $\varphi : 2^X \to \{0, 1\}$ and $\psi : 2^Y \to \{0, 1\}$ to represent two CNF formulas, the product $\varphi \cdot \psi$ is the Boolean function corresponding to the conjunction $\varphi \land \psi$.

\begin{definition} [Projection]
	\label{def:projection}
    Let $X$ be a set of variables and $x\in X$. The projection of $f:2^X \to \mathbb{R}$ w.r.t. $x$ is the function $\Sigma_x f:2^{X\setminus \{x\}} \to \mathbb{R}$ defined for all $\tau \in 2^{X\setminus \{x\}}$ by $(\Sigma_x f)(\tau) :=  f(\tau \cup \{x\}) + f(\tau \cup \{\overline{x}\})$ .
\end{definition}

Note that {\it projection} is commutative, i.e., that $\Sigma_x\Sigma_y f = \Sigma_y\Sigma_x f$ for all variables $x, y \in X$ and functions $f:2^X \to \mathbb{R}$. 
Given a set $X = \{x_1, x_2, \dots, x_n\}$, we define $\Sigma_X f = \Sigma_{x_1} \Sigma_{x_2} \dots \Sigma_{x_n}f$ and $\Sigma_{\varnothing} f = f$. 

The equation in Definition \ref{def:WMC} can be expressed in an alternative form as $W(\psi) = (\Sigma_{X} (\psi \cdot W))(\varnothing)$, whose computation relies on the operations of {\it product} and {\it projection}.
Dudek {\it et al.} showed that a special type of projection, called early projection, can accelerate WMC solving~\cite{DBLP:conf/aaai/DudekPV20}.

\begin{theorem}[Early projection]
        Let $X$ and $Y$ be sets of variables. Given $f:2^X \to \mathbb{R}$ and $g:2^Y \to \mathbb{R}$, if $x\in X \setminus Y$, then $\Sigma_x(f\cdot g) = (\Sigma_x f)\cdot g$ .      
\end{theorem}

The {\it early projection} has critical applications in many fields, such as database-query optimization~\cite{DBLP:journals/jcss/KolaitisV00}, 
satisfiability solving~\cite{DBLP:journals/jar/PanV05}.
 and weighted model counting ~\cite{DBLP:conf/aaai/DudekPV20}.

\subsection{Pseudo-Boolean Constraints}
\label{sct:PBConstraints}

A Pseudo-Boolean (PB) constraint is of the form $\sum_{i=1}^n a_i l_i \vartriangleright b$, where the $a_i$ and $b$ are integers, literal $l_i$ is a Boolean variable $x_i$ or its negation $\overline{x_i}$, i.e., $\overline{l_i} = 1 - l_i$,  
and the relation operator $\vartriangleright \in \{<, \leq, =, \geq, >\}$.
The right side of the constraint $b$ is often called the degree of the constraint.
PB constraints are more expressive than clauses.
Actually, each clause is a special PB constraint in which degree $b$ and all coefficients $a_i$ are $1$, and $\vartriangleright$ is $\geq$. 

A PB formula $\varphi$ is a conjunction of PB constraints. 
$\varphi$ can be represented as a Boolean function $\varphi : 2^{X} \to \{0, 1\}$. 
A complete truth assignment is a mapping that assigns to each variable in the set $X$ either $0$ or $1$.
In PB, a constraint is satisfied when the equality holds. 
Therefore we can regard the PB formulas as a mapping from truth assignment to satisfiability.
We use $\gamma : 2^{X_\gamma} \to \{0, 1\}$ to represent a constraint $\gamma \in \varphi$, where $X_\gamma\subseteq X$ is the set of variables appearing in $\gamma$.

There are many efficient methods for encoding PB constraints into CNF so that we can directly use the tools on CNF to solve the reasoning problems on PB. 
E{\'{e}}n {\it et al.} describe and evaluate three different techniques, translating by BDDs, Adder Network, and Sorting Network~\cite{DBLP:journals/jsat/EenS06}.
There are many studies that have made transformations for a special type of PB constraints, e.g., at-most-one constraints and PBMod-constraints~\cite{DBLP:conf/sara/AavaniMT13,DBLP:journals/ai/BofillCNSUV22}.





\subsection{Algebraic Decision Diagrams}

Algebraic Decision Diagram (ADD) is a compact representation of a function as a directed acyclic graph~\cite{DBLP:journals/fmsd/BaharFGHMPS97}. It is an extension of the Binary Decision Diagram (BDD) that represents functions mapping each state to a finite set of values, e.g., integers representing costs.

We consider an ADD $\mathcal{A}$ over a set of Boolean variables $X$ and an arbitrary set $S$.
$\mathcal{A}$ is a rooted directed acyclic graph where each node of $\mathcal{A}$ is an element in either $X$ or $S$.
Every terminal node of $\mathcal{A}$ is labeled with an element of $S$, and every non-terminal node of $\mathcal{A}$ is labeled with an element of $X$ and has two out-going edges labeled $0$ and $1$ ({\it 0-edge} and {\it 1-edge} for short).
For the directed path from the root to the terminal node in $\mathcal{A}$, the element represented by the non-terminal node on the path will appear no more than once.



\section{WMC on PB Constraints}
\label{sec:WMConPB}

In this section, we introduce the algorithmic framework for weighted model counting on PB constraints. 
We first define the WMC task on PB as follows:

\begin{definition}[Model Counting on PB Constraints]
	\label{def:WMConPB}
	Let $\varphi$ be a PB formula over variables set $X$, and let $W:2^{X} \to \mathbb{R}$ be an arbitrary function. The weighted model count on $\varphi$ w.r.t. $W$ is defined as $W(\varphi) = \sum_{\tau \in 2^{X}} \varphi(\tau) \cdot W(\tau)$	 .
\end{definition}

Obviously, the {\it product} operation can be used to represent the entire formula by constraints, i.e., $\varphi = \prod_{\gamma \in \varphi} \gamma$.
Then the equation in Definition \ref{def:WMConPB} can be rewritten as $W(\varphi) = (\Sigma_X(\varphi \cdot W))(\varnothing)$.
Just like ADDMC, our WMC algorithm uses ADDs to perform {\it products} and {\it projections}.
To facilitate the transformation from PB constraints to ADDs, we normalize the PB constraints.
It is widely recognized that the PB constraint can be normalized using only one operator, such as $\geq$.
In this context, a constraint $\sum_{i=1}^n a_i l_i=b$ is split into two constraints $\sum_{i=1}^n a_il_i\geq b$ and $\sum_{i=1}^n a_i(\overline{l_i} - 1)\geq -b$.
However, when computing WMC, we need to perform {\it product} between constraints, and therefore the split incurs an additional cost.
To address this issue, we relax the standard normalized PB constraint to admit $=$ in addition to $\geq$.

\begin{definition}[Relaxed normalized PB constraint]
	Given a PB constraint $\sum_{i=1}^n a_il_i \vartriangleright b$, it is relaxed normalized iff each coefficient $a_i$ is non-negative integer and operator $\vartriangleright$ is $\geq$ or $=$. 
\end{definition}

We can transform each PB constraint into a relaxed normalized PB constraint as follows.
First, we transform $\vartriangleright$ into $\geq$ except for $=$. 
That is, we multiply both sides of the inequality by $-1$ to transform the operators $<$ (resp. $\leq$)  into $>$ (resp. $\geq$), and then we can replace $> b$ by $\geq b+1$. 
Second, we can replace $-al$ by $a\overline{l}-a$ to ensure every coefficient is a non-negative integer.
Finally, if a relaxed normalized PB constraint has a negative degree, it cannot be satisfied and therefore the weighted model count is zero.
Hereafter, we assume each PB constraint is relaxed normalized unless otherwise stated.

\subsection{Dynamic Programming Framework}

The focus of dynamic programming is to decompose complex problems into small sub-problems to solve and combine the solutions to get the answer.
We extend the algorithmic framework of ADDMC~\cite{DBLP:conf/aaai/DudekPV20} to solve the WMC on PB.
Our algorithm is presented in Algorithm \ref{al:PBCounter}, and we will discuss our preprocessing techniques in Section \ref{sec:preTech}~(line 1).

The input $\prec$ is called diagram variable order and is used to build ordered ADDs to support tractable {\it products} and {\it projections}.
It determines the order of variable occurrence along the path from the root to the terminal node in ADDs and greatly influences the size of the diagram.

Another input $\rho$ is injection $X \to \mathbb{Z}^+$, where $X = \texttt{Vars}(\varphi)$ represents all variables appearing in formula $\varphi$. 
$\rho$ is used to partition PB constraints into several disjoint parts $\{\Gamma_i\}_{i=1}^{m}$, called clusters. 
According to these clusters, we get the partition $\{X_i\}_{i=1}^m$ of the variable set $X$.
Variables already placed in $X_i$ will not be placed in $X_1, \dots X_{i-1}$. 
This property allows us to {\it early project} the variables in $X_i$ (line 11), because they will not appear later.

\begin{algorithm}
    \caption{\PBCounter{}$(\varphi, W, \prec, \rho)$ : WMC on PB Constraints}
    \label{al:PBCounter}
    \KwIn{nonempty PB formula $\varphi$, literal-weight function $W$ over $X = \texttt{Vars}(\varphi)$, diagram variable order $\prec$, cluster variable order $\rho$ } 
    \KwOut{ $W(\varphi)$ weighted PB count of $\varphi$ w.r.t. $W$}
    
    $\varphi \gets \texttt{preprocess}(\varphi)$ \\
    $m \gets \max_{x\in Vars(\varphi)} \rho(x)$ \\
    $\varphi \gets \texttt{sort}(\varphi, \prec)$
    
    \For{$i = m, m-1, \dots 1$}{
        $\Gamma_i \gets \{\gamma \in \varphi : \texttt{constraintRank}(\gamma, \rho) = i\}$ \\
        $\kappa_i \gets \{\texttt{constructEqADD}(\gamma, 1, b) : \gamma \in \Gamma_i, \vartriangleright \text{is} =\} \cup \{\texttt{constructGEqADD}(\gamma,1,b) : \gamma \in \Gamma_i, \vartriangleright \text{is} \geq\}$ \\
        $X_i \gets \texttt{Vars}(\Gamma_i)\setminus \bigcup_{j=i+1}^m \texttt{Vars}(\Gamma_j)$ \\
    }
    \For {$i = 1, 2, \dots m$} {
        \If {$\kappa_i \neq \varnothing$}{
            $\mathcal{A}_i \gets \prod_{D\in \kappa_i} D$ \;
            \lFor {$x \in X_i$} {$\mathcal{A}_i \gets \Sigma_x(\mathcal{A}_i \cdot W_x)$}
            \If{$i < m$}{
                $j \gets \texttt{chooseCluster}(\mathcal{A}_i, i)$ \\
                $\kappa_j \gets \kappa_j \cup \{\mathcal{A}_i\}$ \\
            }
        }
    }
    \Return $\mathcal{A}_m(\varnothing) $
\end{algorithm}

We use $\kappa_i$ for each cluster to represent the set of ADDs in the cluster $\Gamma_i$. We {\it product} ADDs in $\kappa_i$ (line 10), do {\it project}, and insert the result ADD $\mathcal{A}_i$ into another cluster (line 14).

Heuristics are essential parts of Algorithm \ref{al:PBCounter}. 
We implement all the heuristics used in ADDMC for PB constraints and choose \textbf{MCS} (maximum-cardinality search~\cite{DBLP:journals/siamcomp/TarjanY84}) and \textbf{LexP} (lexicographic search for perfect orders~\cite{DBLP:journals/endm/KosterBH01}) as the diagram variable order ($\prec$) and cluster variable order ($\rho$), respectively, and \textbf{BM-Tree} (Bouquet's Method with tree data structure~\cite{DBLP:conf/aaai/DudekPV20}) as the cluster heuristic for the function \texttt{constraintRank} and \texttt{chooseCluster}.
This combination performed best in the experiment by ADDMC.

\subsection{Construct ADDs for PB Constraints}
We show how \PBCounter{} constructs ADDs to represent PB constraints (line 6 in Algorithm \ref{al:PBCounter}).
We use two different construction methods, described in Algorithms \ref{alg:eqADD}--\ref{alg:geqADD}, for the PB constraints with operators $=$ and $\geq$, respectively.

\begin{algorithm}[h]
    \setstretch{1.1}
    \caption{\texttt{constructEqADD($\gamma, j, k$)} : Construct ADD for sub-constraint $\gamma':\sum_{i=j}^n a_il_i = k$}
    \label{alg:eqADD}
    \KwIn{A PB constraint $\gamma: \sum_{i=1}^n a_il_i = b$, integers $1\leq j \leq n + 1, k$}
    \KwOut{$\mathcal{A}$: ADD represent sub-constraint $\gamma' : \sum_{i=j}^n a_i l_i = k$}
        
    \lIf{$k < 0$} {\textbf{return} $TerminalNode(0)$ \text{// UNSAT for all assignments}}
    \If{$j = n + 1$} {
    	\lIf{$k = 0$} {\Return $TerminalNode(1)$ \text{// Find a SAT assignment}}
    	\lElse{\Return $TerminalNode(0)$}
    }
    $\mathcal{A} \gets \texttt{findEqADD}(\gamma, j, k)$ \\
    \lIf{$\mathcal{A} \neq \varnothing$}{\Return $\mathcal{A}$}
    $\mathcal{A}_f \gets \texttt{constructEqADD}(\gamma, j+1, k)$ \\
    \tcp{$\mathcal{A}_f$ represent $\sum_{i=j+1}^n a_i l_i = k \land \overline{l_j}$}
    $\mathcal{A}_t \gets \texttt{constructEqADD}(\gamma, j+1, k-a_j)$ \\
    \tcp {$\mathcal{A}_t$ represent $(\sum_{i=j+1}^n a_i l_i = k - a_j) \land l_j$}
	
	\lIf{$l_j$ is positive}{$\mathcal{A} \gets \texttt{Ite}(x_j, \mathcal{A}_t, \mathcal{A}_f)$}
	\lElse{$\mathcal{A} \gets \texttt{Ite}(x_j, \mathcal{A}_f, \mathcal{A}_t)$}
    \texttt{storeEqADD}$(\gamma, j, k, \mathcal{A})$ \\
    \Return $\mathcal{A}$
\end{algorithm}

In Algorithm \ref{alg:eqADD}, we provide a recursive function for constructing ADD for constraint $\sum_{i=1}^n a_il_i = b$.
It divides the current constraint into two sub-constraints (lines 7-8) and solves them recursively, then constructs the ADD of the current constraint (lines 9-10).
\texttt{Ite}$(x_j, \mathcal{A}_t, \mathcal{A}_f)$, known as {\it If-Then-Else}, is an operation of ADDs, which returns an ADD rooted at $x_j$, {\it 1-edge} linked $\mathcal{A}_t$ and {\it 0-edge} linked $\mathcal{A}_f$.
We use \texttt{storeEqADD} to store constructed sub-constraints with the same parameter $k$ (line 11), avoiding duplicate constructions.
If the same ADD has been constructed before, we can return it directly (lines 5-6).

For each PB constraint with positive literals and $\le$, Ab{\'{\i}}o {\it et al.} proposed a linear algorithm for converting it into a reduced ordered BDD~\cite{DBLP:journals/jair/AbioNORM12}.
We modify this algorithm to make it be able to deal with negative literals and output ADDs, which is an extension of BDDs.
For a PB constraint $\sum_{i=1}^n a_i l_i \geq b$, we can partition the interval  $(-\infty, \infty)$ into multiple sub-intervals such that when $b$ belongs to the same sub-interval, their corresponding ADDs are the same.
We use Example \ref{exp:intervalADDs} to explain this point further.

\begin{figure}[h]
    \centering
    \includegraphics[width=\linewidth]{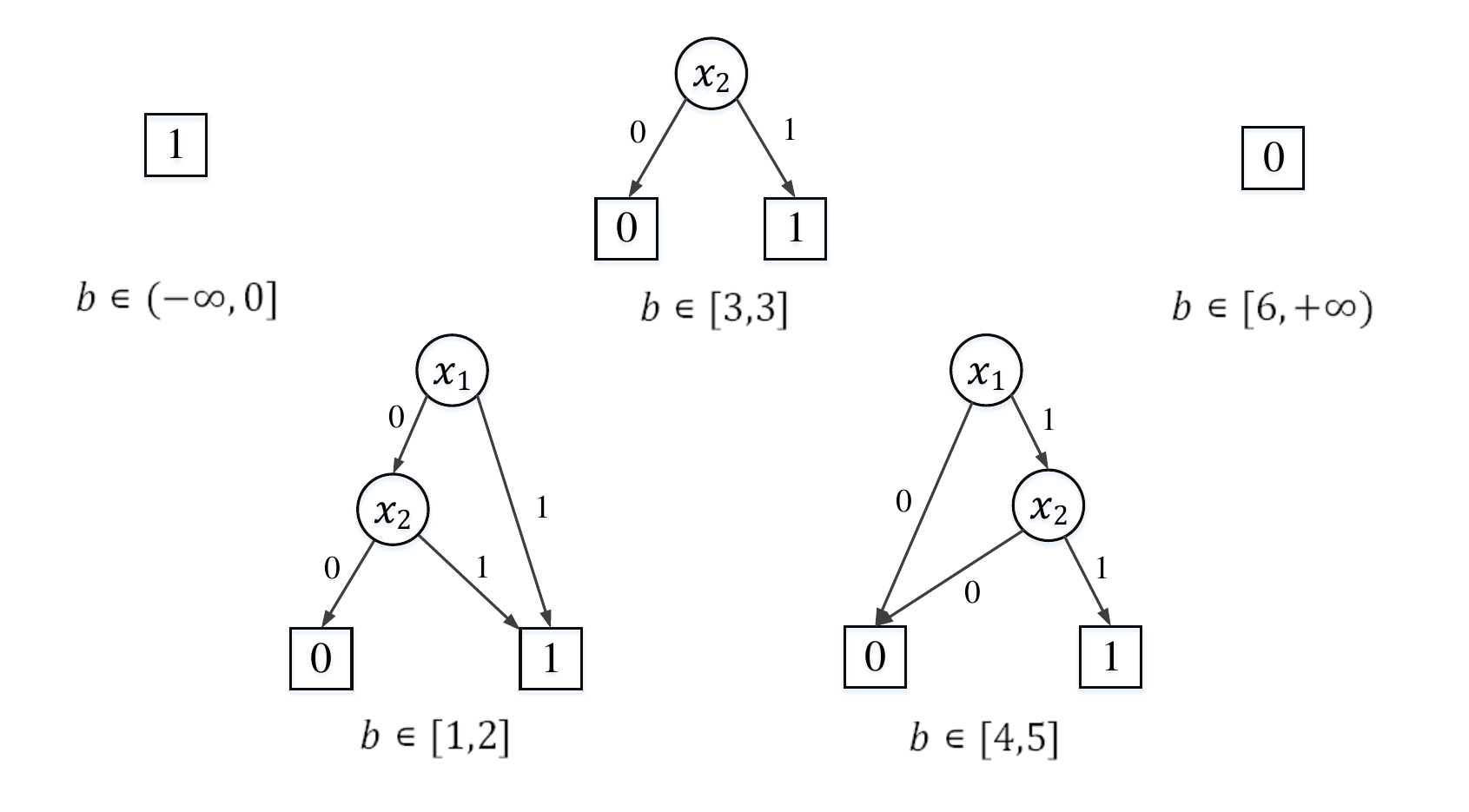}
    \caption{The five possible ADDs of $2x_1 + 3x_2 \geq b$ over $x_1 \prec x_2$.}
    \label{fig:Example2}
\end{figure}

\begin{example}
	The constraint $2x_1 + 3x_2 \geq b$ with the order $x_1 \prec x_2$ can correspond to five possible ADDs when $b$ has different values from five intervals, as shown in Figure \ref{fig:Example2}.
	If $b\in (-\infty, 0]$, every possible assignment over $\{x_1, x_2\}$ satisfies the constraint, and we obtain a single ADD with only one terminal node labeled with $1$.
	Similarly, $2x_1+3x_2 \geq 4$ and $2x_1 + 3x_2 \geq 5$ will be both satisfied when $x_1 = x_2 = true$.
	\label{exp:intervalADDs}
\end{example}

The modified algorithm is presented in Algorithm \ref{alg:geqADD}. 
Consider a PB constraint $\sum_{i=1}^n a_il_i \geq b$ with an order $x_1 \prec x_2 \prec \dots \prec x_n$. 
When we construct an ADD $\mathcal{A}$ for sub-constraint $\sum_{i=j}^n a_il_i \geq k$, we store this ADD and the corresponding interval $[\alpha, \beta]$.
When we compute a new sub-constraint $\sum_{i=j}^n a_il_i \geq k'$ with $k' \in [\alpha, \beta]$, we can return the stored ADD and interval (lines 3--4).
Otherwise, we need to recursively construct ADDs for sub-constraints (lines 5--6). With the outputs of recursive calls, we can compute the resulting ADD and the corresponding interval (lines 7--9).

\begin{algorithm}[h]
    \setstretch{1.1}
    \caption{\texttt{constructGEqADD($\gamma, j, k$)} : Construct ADD for sub-constraint $\gamma':\sum_{i=j}^n a_il_i \geq k$}
    \label{alg:geqADD}
    \KwIn{A PB constraint $\gamma: \sum_{i=1}^n a_il_i \geq b$, integers $1\leq j \leq n + 1, k$}
    \KwOut{$[\alpha, \beta]$: interval contains $k$, \newline $\mathcal{A}$: ADD represent sub-constraint $\gamma' : \sum_{i=j}^n a_i l_i \geq [\alpha, \beta]$}
    
    \If{$k < 0$} {\Return $([-\infty, -1], TerminalNode(0))$}
    \If{$k \geq \sum_{i=j}^n a_i$} {\Return $([\sum_{i=j}^n a_i, \infty], TerminalNode(1))$}
    
    $([\alpha, \beta],\mathcal{A}) \gets \texttt{findGeqADD}(\gamma, j, k)$ \\
    \tcp{Find a stored ADD for $\gamma'$ where $k\in [\alpha, \beta]$}
    \lIf{$\mathcal{A} \neq \varnothing$} { \Return $([\alpha, \beta], \mathcal{A}$)}

    $([\alpha_f, \beta_f], \mathcal{A}_f) \gets \texttt{constructGEqADD}(\gamma, j+1, k)$ \\
    $([\alpha_t, \beta_t], \mathcal{A}_t) \gets \texttt{constructGEqADD}(\gamma, j+1, k - a_i)$ \\
	\lIf{$l_j$ is positive}{$\mathcal{A} \gets \texttt{Ite}(x_j, \mathcal{A}_t, \mathcal{A}_f)$}
	\lElse{$\mathcal{A} \gets \texttt{Ite}(x_j, \mathcal{A}_f, \mathcal{A}_t)$}
	$[\alpha, \beta] \gets [\alpha_f, \beta_f] \cap [\alpha_t + a_j, \beta_t + a_j]$ \\ 
   
    $\texttt{storeGeqADD}(\gamma, j, [\alpha, \beta], \mathcal{A})$ \\
    \Return $([\alpha, \beta], \mathcal{A})$
\end{algorithm}

Note that Algorithm \ref{alg:geqADD} returns a tuple of an interval $[\alpha, \beta]$ and ADD $\mathcal{A}$.
That means every $k \in [\alpha, \beta]$ has same ADD $\mathcal{A}$ for sub-constraint $\gamma': \sum_{i=j}^n a_il_i \geq k$.
In line 5 of Algorithm \ref{al:PBCounter}, we need to ignore intervals and only keep ADDs in $\kappa_i$.

\section{Preprocessing Techniques for WMC on PB}
\label{sec:preTech}

Efficient preprocessing techniques are an essential part of model counting solvers.
Solvers like SharpSAT-TD and ExactMC have built-in preprocessors~\cite{korhonen2021integrating,DBLP:conf/aaai/LaiMY21}, and some preprocessors designed explicitly for Model Counting, such as {\it pmc} and {\it B+E}~\cite{DBLP:journals/ai/LagniezLM20}, have also demonstrated the significance of preprocessing.
However, there are no dedicated pre-processing techniques for WMC on PB currently.
We first applied two preprocessing techniques to pseudo-Boolean formulas as our built-in preprocessing methods.


The first technique is based on the backbone.
The backbone of a formula $\varphi$ is the set of all literals which are implied by $\varphi$ when $\varphi$ is satisfiable (variables that have the same value in all solutions) and is the empty set otherwise~\cite{DBLP:conf/aaai/LagniezM14}.
Algorithm \ref{alg:backBone} simplifies the PB formula $\varphi$ by identifying backbone variables and propagating unit literals on these backbone variables.
In our implementation, we use RoundingSAT~\cite{Elffers:Nordstrom:18} to solve a PB formula and return a solution that satisfies the formula (lines 2, 4). If the formula is unsatisfiable, an empty set is returned.
Similar to CNF, when a literal is determined to be true, the unit propagation on PB removes the literal and its negation from all other constraints.
It simultaneously eliminates the term ($al$ or $a\overline{l}$) from both sides of the inequality.
While the constraint is determined to be satisfiable, it can be deleted; if it is determined to be unsatisfiable, return $\bot$.

\begin{algorithm}[t]
    \setstretch{1.1}
    \caption{\texttt{addBackbone}($\varphi$)}
    \label{alg:backBone}
    \KwIn{A PB formula $\varphi$}
    \KwOut{A PB formula equivalent to $\varphi$}
    
    $\mathcal{B} \gets \varnothing$ \\
    $\mathcal{I} \gets \texttt{solve}(\varphi)$
    
    \While{$\exists l \in \mathcal{I}$ s.t. $l \notin \mathcal{B}$} {
    	$\mathcal{I}' \gets \texttt{solve}(\varphi \land \overline{l}\,)$ \\
    	\If{$\mathcal{I}'$ is $\varnothing$} {
    		$\mathcal{B} \gets \mathcal{B} \cup \{l\}$ \\
    		$\varphi \gets \texttt{unitPropagate}(\varphi, l\,) \land \{l = 1\}$
    	}
    	\lElse {$\mathcal{I} \gets \mathcal{I} \cap \mathcal{I}'$}
    }
    
    \Return $\varphi$
\end{algorithm}

Since \PBCounter{} requires the construction and operations of ADDs for each constraint, we propose the second preprocessing technique named \texttt{deleteConstraint}.
It simplifies the formula by removing constraints $\gamma : [\varphi \setminus \gamma ] \models \gamma$ representing in Algorithm \ref{alg:delCons}.
To verify $[\varphi \setminus \gamma ] \models \gamma$, which is equivalent to $[\varphi \setminus \gamma ] \land \overline{\gamma} \models \bot$, we transform $\gamma$ into its corresponding CNF formula, denoted as $c_1 \land c_2 \land \dots \land  c_m$.
Then, we can proceed to validate $[\varphi \setminus \gamma ] \land (\overline{c_1} \lor \overline{c_2}  \lor \cdots \lor \overline{c_m} ) \models \bot$.
Using the algorithm presented in Section \ref{sec:WMConPB}, we can construct a ADD $\mathcal{A}$ for constraint $\gamma$ (line 4).
Each non-terminal node in $\mathcal{A}$ is labeled with a variable from $X$, and edges from each node represent literals (1-edge corresponds to positive literal, 0-edge corresponds to negative literal).
That means all $\overline{c_1}, \overline{c_2}, \dots, \overline{c_m}$ are unsatisfiable assignments for $\gamma$, and correspond to the path from the root node to a terminal node labeled with 0 in $\mathcal{A}$ (line 5).

\begin{example}
	The PB formula $\varphi$ includes three constraints, $\gamma_1 : 3x_1 + 4x_2 \geq 4$, $\gamma_2 : 3x_1 + x_3 + x_4 \geq 4$, and $\gamma_3 : 3x_2 + x_3 + x_4 \geq 4$.
	According to Figure \ref{fig:Example2}, the ADD of $\gamma_1$ has two paths, $\overline{x_1}$ and $x_1 \land \overline{x_2}$, to terminal node 0.
	Therefore, we get $\overline{c_1} = \overline{x_1}$ and $\overline{c_2} = x_1 \land \overline{x_2}$.
	Since $[\gamma_2\land \gamma_3]\land \overline{c_1} \models \bot$ and $[\gamma_2\land \gamma_3]\land \overline{c_2} \models \bot$, we can delete $\gamma_1$ from $\varphi$.
	\label{exp:delCons}
\end{example}

To avoid the exponential explosion of ADD size, we do not attempt to delete constraints that involve more than 20 literals (line 2).
We use \texttt{unitPropagation} to propagate all literals in unsatisfiable assignment $\overline{c_i}$ (line 8).
If every $c_i$ satisfies $[\varphi \setminus \gamma ] \land \overline{c_i} \models \bot$, $\gamma$ is removed.

\begin{algorithm}[h]
    \setstretch{1.1}
    \caption{\texttt{deleteConstraint}($\varphi$)}
    \label{alg:delCons}
    \KwIn{A PB formula $\varphi$}
    \KwOut{A PB formula equivalent to $\varphi$}
    
    \ForEach{ $\gamma \in \varphi$ }{
    	\lIf{$\texttt{notConsider} (\gamma)$}{\textbf{continue}}
    	$\varphi \gets \varphi \setminus \{\gamma \}$ \\
		$\mathcal{A} = \texttt{constructADD}(\gamma)$ \\
		\ForEach{path from root to $TerminalNode(0)$ in $\mathcal{A}$}{
			$\varphi' \gets \varphi$ \\
			\ForEach{literal l on path}{
				$\varphi' \gets \texttt{unitPropagate}(\varphi', l\,)$
			}
			\If{$\varphi'$ is not $\bot$} {
				$\varphi \gets \varphi \cup \{\gamma\}$ \\
				\textbf{break} \\
			}
		}   	
    }
    
    \Return $\varphi$
\end{algorithm}

To achieve a more efficient implementation of Algorithm \ref{alg:delCons}, we traverse all paths in $\mathcal{A}$ from the root to terminal nodes using a depth-first search.
Based on the edges traversed along the path, we perform unit propagation on the literals it represented.
When we reach the terminal node labeled 0 without encountering any conflicts, it implies there exists a $\overline{c_i}$ such that $[\varphi \setminus \gamma ] \land \overline{c_i} \not\models \bot$.
Therefore, the corresponding clause $\gamma$ can not be removed~(line 10).
We can directly backtrack when a conflict occurs on non-terminal nodes.

\section{Experimental Results}
\label{sec:ExpRes}

In this section, we conducted experiments to evaluate the performance of our solver \PBCounter{}. We implemented \PBCounter{} in C++. All experiments were run on a computer with Intel(R) Xeon(R) W-2133 CPU @ 3.60GHz and 64GB RAM.

\subsection{Experiment Settings}

As \PBCounter{} is the first lazy WMC solver on PB constraints, we compare it with eager WMC solvers, an ILP counter, and a PB enumerator.
We implemented two WMC encoding methods in C++ and combined them with four WMC counters on CNF. 
In other words, we have eight competitors for solving the WMC problem based on CNF, and two other approaches based on PB constraints.
We ran each combination and \PBCounter{} on each benchmark using a single core, 16 GB of memory, and 1800 seconds of the time limit.

\noindent \textbf{Competitive Solvers}

The four WMC solvers on CNF formulas include SharpSAT-TD, ExactMC, D4, and ADDMC. 
SharpSAT-TD is the winner of the Model Counting Competition 2021 - 2023 in the weighted track. It is based on solver SharpSAT and combines the technology of tree decomposition \cite{DBLP:conf/sat/Thurley06,korhonen2021integrating}.
ExactMC is based on CCDD (a generalization of Decision-DNNF), which can capture literal equivalences~\cite{DBLP:conf/aaai/LaiMY21}. 
After compiling the instance into CCDD, ExactMC can compute model counts in linear time.
D4 is a Decision-DNNF compiler that can calculate model counts by knowledge compilation~\cite{DBLP:conf/ijcai/LagniezM17}.
ADDMC is the winner of the Model Counting Competition 2020 in the weighted track. It is a dynamic programming algorithm using ADD as the main data structure~\cite{DBLP:conf/aaai/DudekPV20}.
We use the decision diagram package {\it Sylvan} to manage ADDs in ADDMC and our solver \PBCounter{}. {\it Sylvan} can interface with the GNU Multiple Precision library to support ADDs with higher-precision numbers~\cite{van2017sylvan}.

The two encoding methods we used are Warners~\cite{DBLP:journals/ipl/Warners98} and ArcCons~\cite{DBLP:journals/jsat/BailleuxBR06}, which are proved counting safe~\cite{DBLP:conf/sat/MorgadoMMM06}.
We implemented the weighted encoders based on these two methods by keeping the original literals' weights and setting the auxiliary literals' weights to 1. This ensures that the weighted model counts are equal.

We also compared two other approaches.
One is to run the PB solver RoundingSAT~\cite{Elffers:Nordstrom:18} to enumerate PB solutions and compute weighted model counts.
The enumeration is achieved by continually adding the negation of the resulting satisfying assignments (called blocking clauses) to the formula to enumerate until the given formula is falsified.
The other one uses ILP counter IntCount~\cite{ge2021decomposition}, which introduces column and row elimination techniques to decompose integer counting problems for linear constraints into independent sub-problems.
Since IntCount cannot solve weighted instances, IntCount runs on an unweighted version of the benchmark.

\noindent \textbf{Benchmarks}

Our benchmarks consist of three parts.
1) We select instances from PB Competition 06, excluding clause-style instances since they can be solved directly with the WMC solver, resulting in 287 instances.
2) We select the benchmarks of Exact Cover used by Junttila and Kaski, consisting of 539 instances from 7 different problem families~\cite{DBLP:conf/cp/JunttilaK10}.
3) According to the generator proposed by Kov{\'a}sznai, we randomly generated 600 instances of Wireless Sensor Network (WSN) problem for six different combinations of three types of sensor and target nodes (5-2, 6-2, 7-3), and two constraint sets (all constraints on and evasive \& moving off)~\cite{kovasznai2018investigations}. One hundred instances were generated for each configuration.

For these instances, we randomly assigned weights $W(x) \in (0,1)$ and $1-W(x)$ to each $x$ and $\overline x$, respectively. 
For SharpSAT-TD, ExactMC, and D4, random weights have no effect on their performance. For ADDMC and \PBCounter{}, random weights may increase the number of terminal nodes in ADDs and therefore may increase the times of operating ADDs.

\subsection{Evaluation}

\noindent \textbf{Experiment on Preprocessing Techniques}

%

Firstly, we compared the impact of the preprocessing techniques mentioned in Section \ref{sec:preTech} on \PBCounter{}.
We used "NoPre" to represent \PBCounter{} without any preprocessing techniques, and "OnlyBb" to represent \PBCounter{} with only the backbone technique (algorithm \ref{alg:backBone}).
Table \ref{tab:preTech} presents the results of the experiment for various benchmarks.
We emphasize that the number of solved benchmarks only counts satisfiable instances. 
Unsatisfiable instances can be judged by pre-calling the PB or SAT solver and have no significance for model counting

\begin{table}[h]
\centering
\begin{footnotesize}
	\renewcommand\arraystretch{1.2}
	\captionsetup{justification=centering}
    \caption{Results of Different Preprocessing Techniques}
    \centering
    \begin{tabularx}{\linewidth}{p{2cm} Z Z Z}
        \hline
        Benchmark (\#) & NoPre & OnlyBb & \PBCounter{} \\
        \cline{2-4}
        \hline
        PB06 (287) &	28  & 37  & \textbf{39}  \\
        ExactCover (539) & 192 & 193 & \textbf{194} \\
        WSN (600) & 387 & 425 & \textbf{483} \\
        Total (1426) & 607 & 655 & \textbf{716} \\
        \hline
    \end{tabularx}
    \label{tab:preTech}
\end{footnotesize}
\end{table}

The results show that the preprocessing techniques of algorithms \ref{alg:backBone} and \ref{alg:delCons} exhibit positive impacts across three different benchmarks.
After incorporating preprocessing techniques into \PBCounter{}, the algorithm was able to solve an additional 109 instances in total. 
Additionally, using only {\it backbone} technique allows \PBCounter{} to solve an additional 48 instances.
Notably, the most significant enhancement was observed in the WSN benchmark.

%

\noindent \textbf{Comparing with WMC Solvers on CNF}
 
\begin{figure*}[h]
    \centering
    \includegraphics[width=0.84\linewidth]{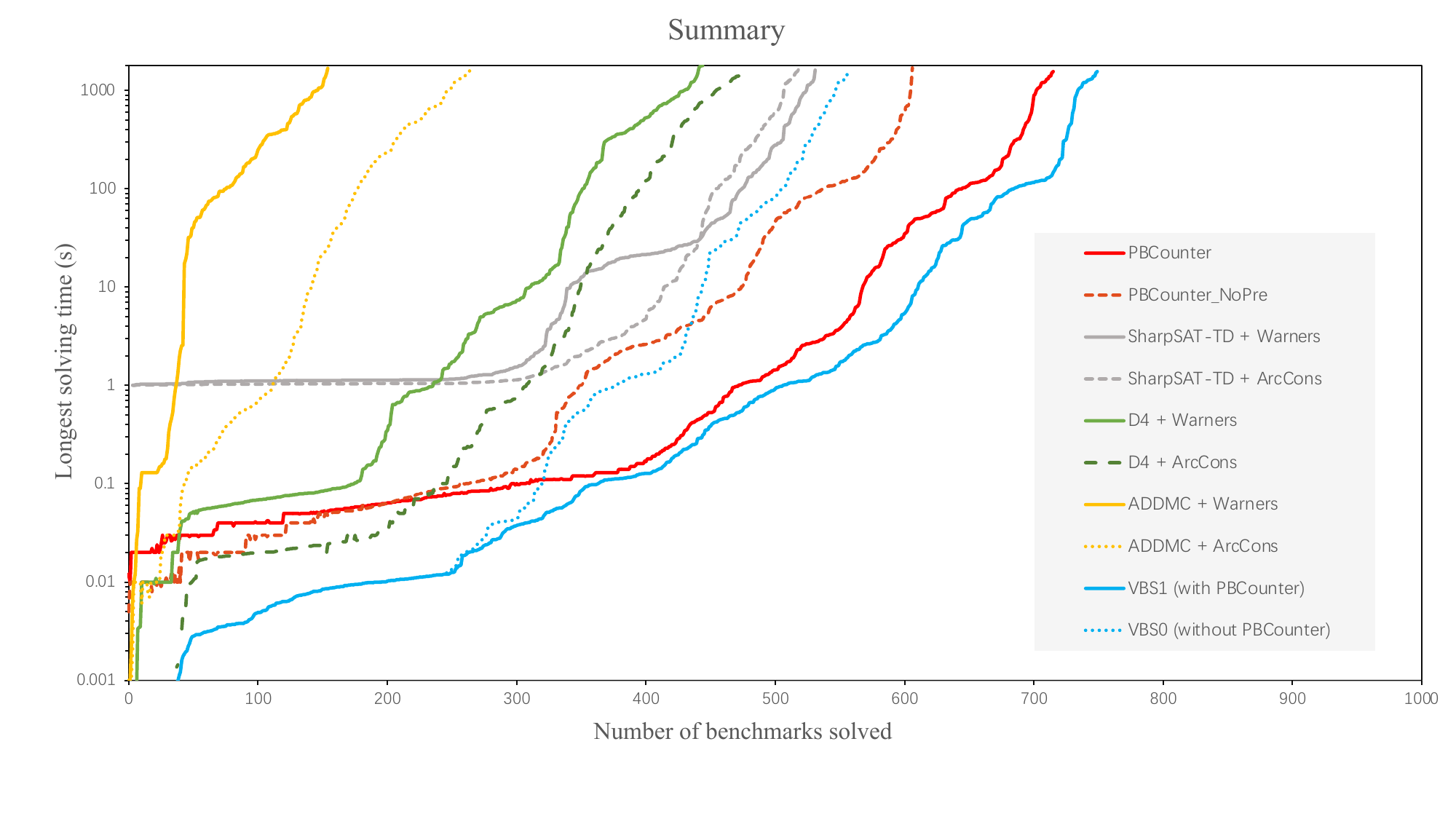}
    \caption{A cactus plot of the numbers of benchmarks solved by seven weighted model counters and two virtual best solvers (VBS1, with \PBCounter{}, and VBS0, without \PBCounter{}).}
    \label{fig:cactusPlot}
\end{figure*}

The performances of \PBCounter{} and other counting combinations are summarized in Table \ref{tab:table1} and Figure \ref{fig:cactusPlot}.
\PBCounter{} is the fastest solver in 327 instances and can uniquely solve 110 instances.
\PBCounter{}\_NoPre turns off preprocessing techniques and solves more instances than CNF-based WMC solvers; it uses the shortest time in 62 instances and uniquely solves 19 instances.

Two virtual best solvers: VBS1 shows the shortest solving time among all counters for each benchmark, while VBS0 shows the shortest time among eight WMC combinations (excluding \PBCounter{}).
Except for \PBCounter{}, SharpSAT-TD solves the most number of instances, while ExactMC has shorter solving times on more instances.


\begin{table}[h]
\centering
\begin{footnotesize}
	\renewcommand\arraystretch{1.2}
	\captionsetup{justification=centering}
    \caption{The number of instances solved by solvers.}
    
    \begin{tabularx}{\linewidth}{p{3.5cm} Z Z Z Z}
        \hline
        \multirow{2}{*}{Solvers}  & \multicolumn{3}{c}{Instances Solved} \\
        \cline{2-4}
        & Unique & Fastest & Total \\
        \hline
        VBS1 (with \PBCounter{})     & - & - & 750 \\
        VBS0 (without \PBCounter{})  & - & - & 557 \\
        \specialrule{0em}{1pt}{1pt}
		\PBCounter{}				 &\textbf{110} &\textbf{327} &\textbf{716} \\
        \PBCounter{}\_Nopre          &\textbf{19} & \textbf{62} & \textbf{607} \\
        \specialrule{0em}{1pt}{1pt}
        SharpSAT-TD + Warner		& 3  &  11	& 531 \\
        SharpSAT-TD + ArcCons	& 0  &  2	& 520 \\
        \specialrule{0em}{1pt}{1pt}
        ExactMC + Warner			& 0  & 3 	& 511 \\
        ExactMC + GenArc			& 0  & 327	& 499 \\
        \specialrule{0em}{1pt}{1pt}
        D4 + Warner			  	& 0  & 0		& 445 \\
        D4 + ArcCons            & 1  & 18	& 480 \\
        \specialrule{0em}{1pt}{1pt}
        ADDMC + Warner			& 0  &  0	& 155 \\
        ADDMC + ArcCons			& 0	 &	0	& 266 \\
        \hline
    \end{tabularx}
    \label{tab:table1}
\end{footnotesize}
\end{table}

For the instances \PBCounter{} uniquely solved, we observed that they often have long constraints. Encoding long constraints into CNF introduces more auxiliary variables and increases the difficulty of solving.
Additionally, if only a few variables appear in multiple long constraints, \PBCounter{} can reduce the size of ADDs by {\it early projection} before {\it product} computations.

\noindent \textbf{Comparing with Different Approaches}

We use "RSBlocking" to denote our implementation of the enumeration solver based on RoundingSAT~\cite{Elffers:Nordstrom:18}.
IntCount is an ILP counter developed by Ge et al~\cite{ge2021decomposition} based on Barvinok~\cite{verdoolaege2007counting}.
Table \ref{tab:2Approach} shows the number of benchmarks solved by \PBCounter{}, RSBlocking, and IntCount.

\begin{table}[h]
\centering
\begin{footnotesize}
	\renewcommand\arraystretch{1.2}
	\captionsetup{justification=centering}
    \caption{Results of Different Approaches}
    \centering
    \begin{tabularx}{\linewidth}{p{2cm} Z Z Z}
        \hline
        Benchmark (\#) & \PBCounter{} & RSBlocking & IntCount \\
        \cline{2-4}
        \hline
        PB06 (287) 		 & \textbf{39}   & 16   & 5  \\
        ExactCover (539) & \textbf{194}  & 178  & 8  \\
        WSN (600) 		 & \textbf{483} & 0    & 23 \\
        Total (1426) 	 & \textbf{716}  & 194  & 36 \\
        \hline
    \end{tabularx}
    \label{tab:2Approach}
\end{footnotesize}
\end{table}

RSBlocking performs well when the number of solutions is small, especially in the ExactCover benchmark.
In fact, with only one solution, the enumeration algorithm only needs to call the PB solver twice.
In the WSN benchmark, the time and space consumption of RSBlocking enumerated solutions becomes unsolvable due to the large satisfiable solution space.
The ILP solver IntCount performs poorly in the three PB benchmarks, even if they are unweighted.
We find that IntCount is more difficult to solve when the number of variables in instances is large and cannot be decomposed.

\section{Conclusion and Future Work}
\label{sec:Conclusion}

In this work, we implemented the first lazy weighted model counter \PBCounter{} on PB constraints with two preprocessing techniques.
Like ADDMC, \PBCounter{} is based on variable elimination and uses dynamic programming techniques.
Our experiments show that \PBCounter{} performed better than the eager state-of-the-art WMC tools; that is, encoding PB constraints into CNF formulas, and then employing ADDMC, SharpSAT-TD, ExactMC, or D4.

The studies \cite{DBLP:conf/aaai/LagniezM14,DBLP:conf/ijcai/LagniezLM16,DBLP:conf/iccad/SoosM22} in model counting on CNF formulas show that the usage of pre-processing is critical for the performance of model counters.
Our experimental results show that the proposed two preprocessing techniques are very effective in improving the performance of PBCounter, which is consistent with the previous studies.
However, on the one hand, the implementation of the two pre-processing techniques is still relatively basic and can be improved further; on the other hand, we can still design more preprocessing techniques. 
For example, there is a lot of exciting work that can transform PB constraints and CNF formulas into each other with the counting-safe property.
Therefore, we can first transform PB constraints into a CNF formula, then pre-process the CNF formula, and finally transform the pre-processed CNF formula into PB constraints.
One future direction of this study is to develop a more effective pre-processor for the WMC task on PB constraints.


\Acknowledgements{This work is supported by NSFC (under Grant No. 61976050, 61972384), Ministry of Education (under Grant No. 2021BCF01002), Jilin Province Natural Science Foundation, and Jilin Provincial Department of Education. Thanks to the editorial board and reviewers for their valuable comments.}

\bibliographystyle{unsrt}
\bibliography{ref,papers}

\Biography{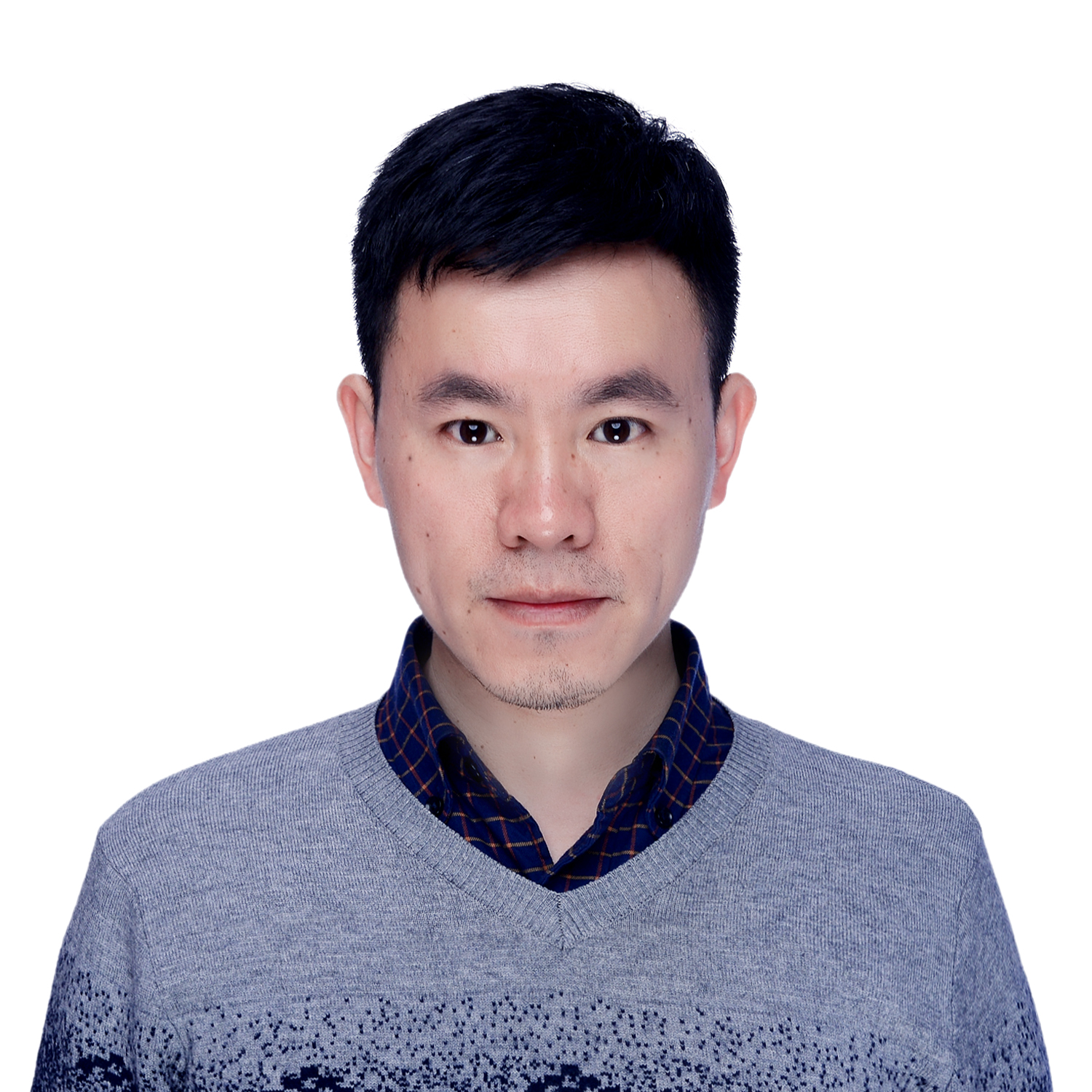}{Yong Lai is an Associate Professor at the College of Computer Science and Technology, Jilin University, China. He received his Ph.D. degree from Jilin University. His research interests include knowledge representation and reasoning, neural-symbolic computation, tractable machine learning, and probabilistic graphical model.}


\Biography{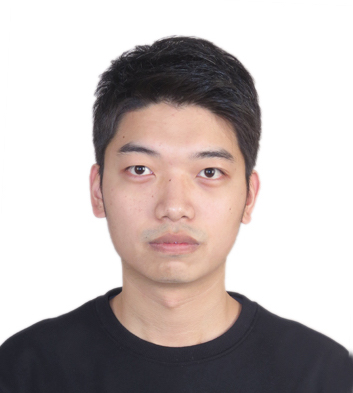} {Zhenghang Xu received a B.S. degree from Computer Science and Information Technology, Northeast Normal University, China in 2020. Currently, he is pursuing a Master's degree at the College of Computer Science and Technology, Jilin University, China. His research interests include model counting and algorithm design.}


\Biography{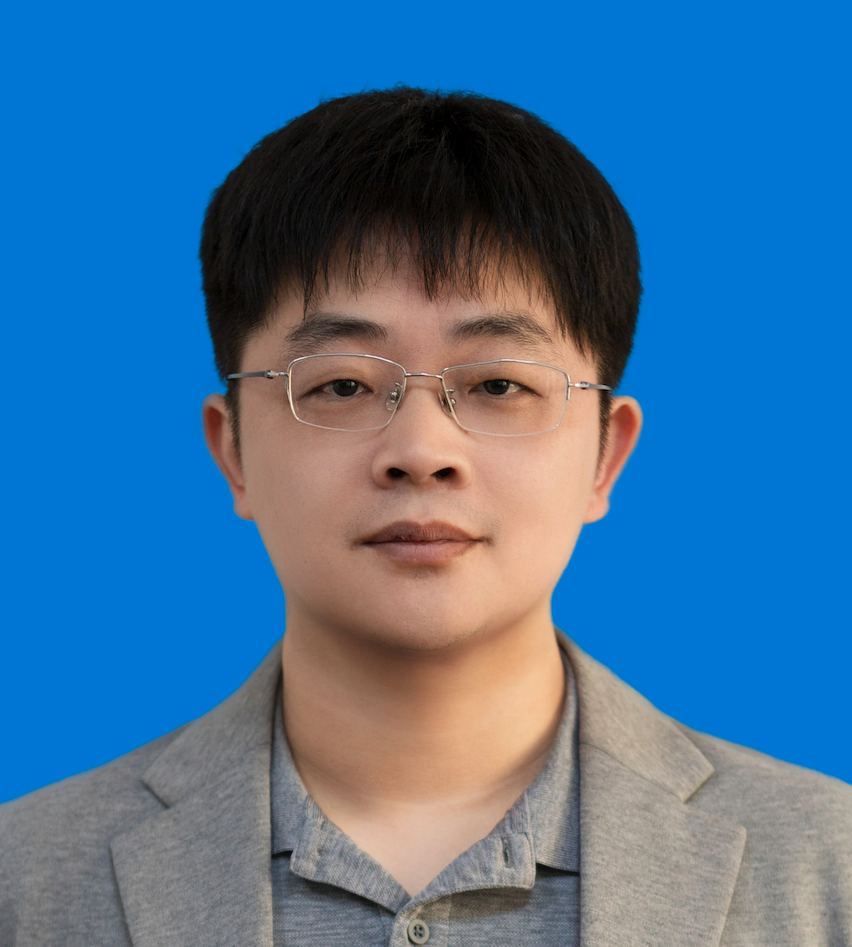}{Minghao Yin is a Professor at the School of Computer Science and Information Technology, Northeast Normal University, China. He received his Ph.D. degree in Computer Software and Theory from Jilin University, China. His research interests include heuristic search, data mining, and combinatorial optimization.}

\end{document}